\documentclass[letterpaper]{article} 
\usepackage{aaai24}  
\usepackage{times}  
\usepackage{helvet}  
\usepackage{courier}  
\usepackage[hyphens]{url}  
\usepackage{graphicx} 
\usepackage{natbib}  
\usepackage{caption} 
\usepackage{algorithm}
\usepackage{algorithmic}
\usepackage{tikz}
\usetikzlibrary{matrix}
\usepackage{amsmath}
\usepackage{tabularx}
\usepackage{makecell}
\usepackage{dblfloatfix}
\usepackage{float}
\usepackage{booktabs}
\usepackage{newfloat}
\usepackage{listings}
\DeclareCaptionStyle{ruled}{labelfont=normalfont,labelsep=colon,strut=off} 
\floatstyle{ruled}
\newfloat{listing}{tb}{lst}{}
\floatname{listing}{Listing}
\title{Combining Deep Learning and Street View Imagery to Map Smallholder Crop Types}
\author {
    Jordi Laguarta Soler\textsuperscript{\rm 1},
    Thomas Friedel\textsuperscript{\rm 2},
    Sherrie Wang\textsuperscript{\rm 1}
}
\affiliations {
    \textsuperscript{\rm 1}Massachusetts Institute of Technology, Cambridge, MA, USA\\
    \textsuperscript{\rm 2}PEAT GmbH, Berlin, Germany\\
    \{laguarta, sherwang\}@mit.edu, thomas@plantix.net
}

\begin{document}

\maketitle

\begin{abstract}
Accurate crop type maps are an essential source of information for monitoring yield progress at scale, projecting global crop production, and planning effective policies. To date, however, crop type maps remain challenging to create in low- and middle-income countries due to a lack of ground truth labels for training machine learning models. Field surveys are the gold standard in terms of accuracy but require an often-prohibitively large amount of time, money, and statistical capacity. 
In recent years, street-level imagery, such as Google Street View, KartaView, and Mapillary, has become available around the world. Such imagery contains rich information about crop types grown at particular locations and times. 
In this work, we develop an automated system to generate crop type ground references using deep learning and Google Street View imagery. The method efficiently curates a set of street-view images containing crop fields, trains a model to predict crop types using either weakly-labeled images from disparate out-of-domain sources or zero-shot labeled street view images with GPT-4V, and combines the predicted labels with remote sensing time series to create a wall-to-wall crop type map. We show that, in Thailand, the resulting country-wide map of rice, cassava, maize, and sugarcane achieves an accuracy of 93\%. We publicly release the first-ever crop type map for all of Thailand 2022 at 10m-resolution with no gaps. To our knowledge, this is the first time a 10m-resolution, multi-crop map has been created for any smallholder country. As the availability of roadside imagery expands, our pipeline provides a way to map crop types at scale around the globe, especially in underserved smallholder regions.
\end{abstract}

\section{Introduction \& Background}

\begin{figure}[]
\centering
\includegraphics[width=0.45\textwidth]{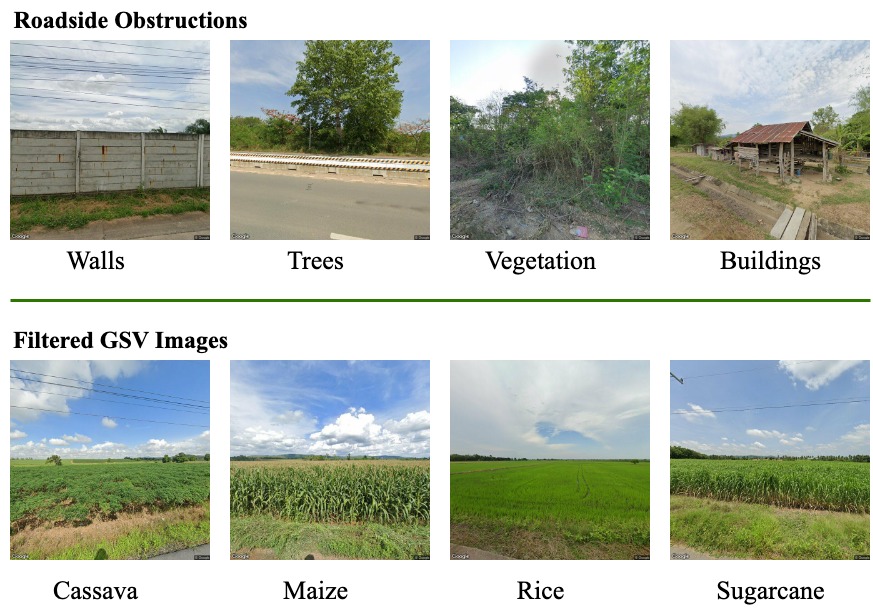}
\caption{Top: Street-view images of roadside occlusions present between the car-mounted camera and fields. Bottom: Street-view images after the automated filtering process for the four major crop types in Thailand.}
\label{fig:obstacles}
\end{figure}

\begin{figure*}[t]
\centering
\includegraphics[width=0.81\textwidth]{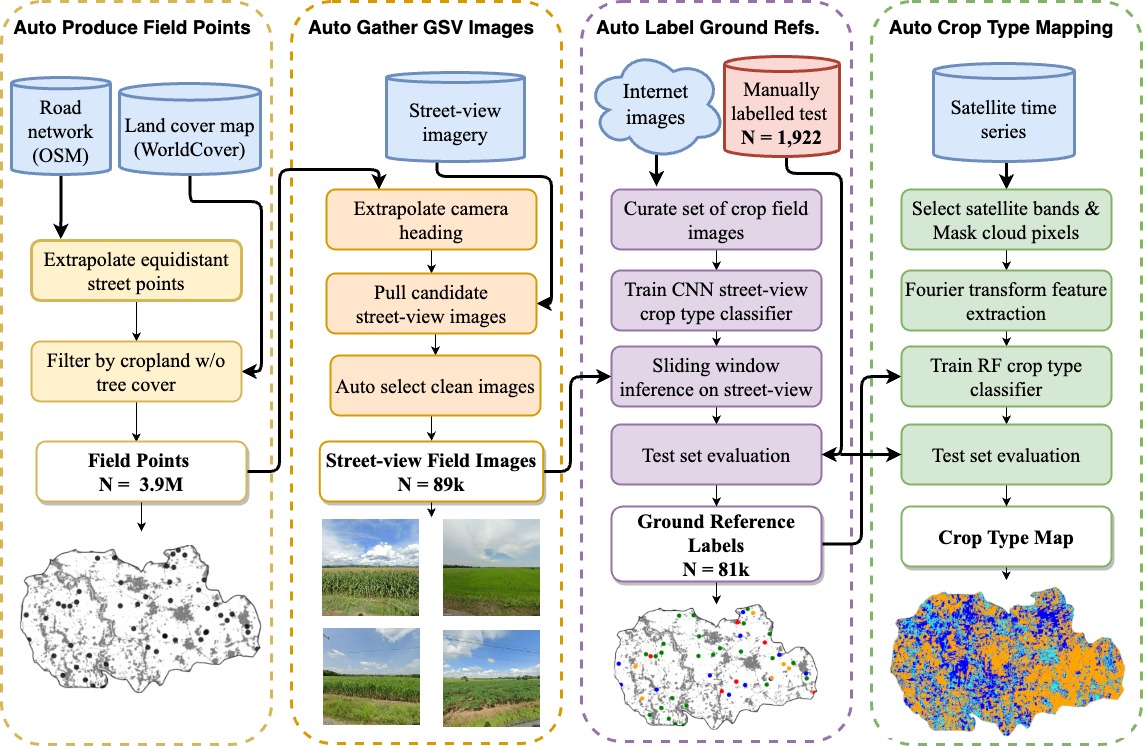}
\caption{Overview of the methods presented in this paper to create a Thailand-wide crop type map. Example field points, ground reference labels, and crop type map are shown for the district of Sawang Ha.}
\label{fig:overview}
\end{figure*}

Ensuring global food security is one of the major challenges we will face this century, especially in the face of a changing climate and a growing population \cite{becker2023crop, rezaei2021crop}. Accurate crop type maps are an essential source of information for monitoring yield progress \cite{becker2023crop}, projecting global crop production \cite{lobell2012influence}, and planning effective policies \cite{haasnoot2013dynamic}. However, only a handful of countries---mostly high-income---have had the budget to collect large-scale ground data and develop crop type maps \cite{han2012cropscape, fisette2013aafc, cantelaube2014registre}. Meanwhile, regions with smallholder farms, which
provide a living for two-thirds of the world's rural population of 3 billion \cite{lowder2016number}
and produce 80\% of the world's food \cite{economic2014state}, continue to lack such maps. 
The majority of smallholder farms are located in middle- and low-income countries, where expensive ground data on crop types remains scarce \cite{tseng2021cropharvest, wang2020mapping, lee2022mapping}. 



To address the high cost of acquiring ground reference labels, the remote sensing and machine learning communities have started to explore non-traditional 
sources of crop type data.
One such source is 
roadside images through services like Google Street View (GSV), Bing Maps StreetSide, Mapillary, KartaView, Tencent Street View, and Baidu Total View. The images are captured by dash cams or panoramic cameras mounted on cars; depending on the service, they are crowdsourced or collected by dedicated fleets. Today, they are low-cost to access, available in almost every country, and updated every few years.

Recent works have used roadside imagery in applications ranging from urban morphology to real estate to air quality prediction \cite{biljecki2021street}. Most relevant to our work, 
\citet{paliyam2021street2sat} 
deployed cameras mounted on the hoods of vehicles in Africa, \citet{wu2021identification} used smartphones mounted on car windows, and
\citet{yan2021exploring} used GSV to create crop type ground reference labels.
Recently, the European Space Agency and WorldCereal Consortium released a crop type map for cereals leveraging GSV to manually create a validation set \cite{van_tricht_kristof_2023_7875105}. 

\begin{figure*}[]
\includegraphics[width=\textwidth]{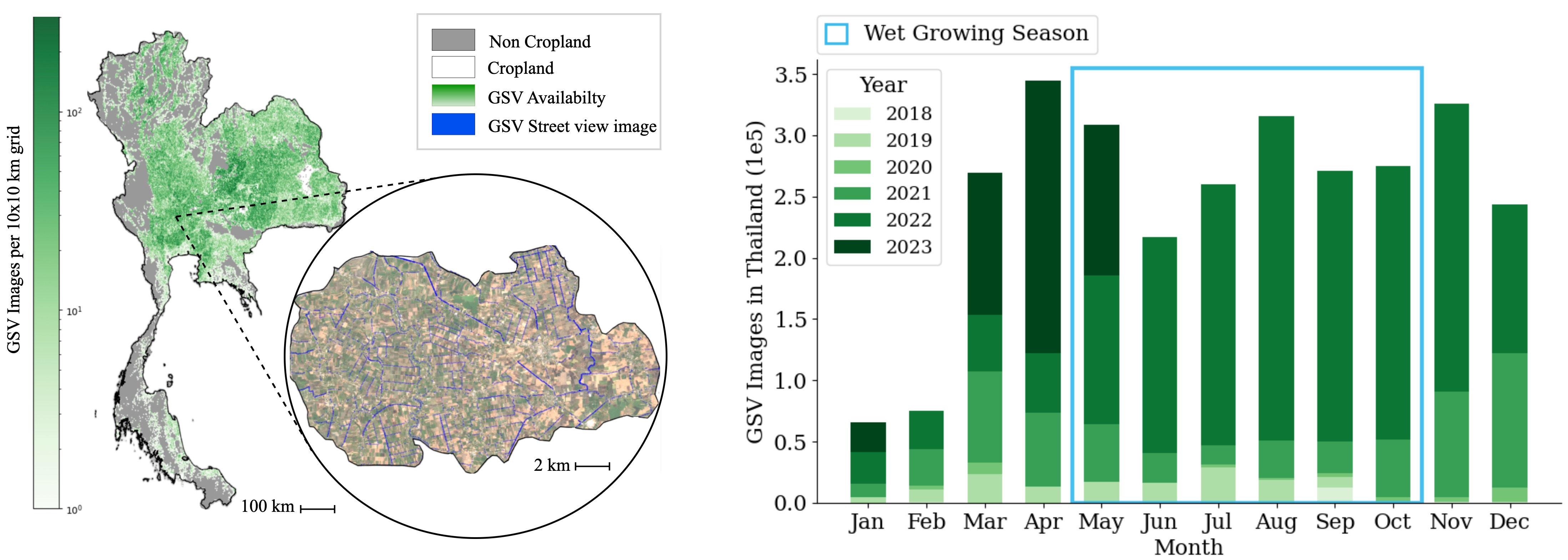}
\caption{Spatial and temporal distribution of Google Street View in Thailand. Left: Hexbin plot of GSV availability across Thailand. The zoomed-in panel shows the location of street-view images overlaid on a satellite basemap in the district of Sawaeng Ha. Right: Availability of street-view images in Thailand by month, with a clear rise in availability since 2022 and a total of over 3 million images. During the wet season (May--October) shown in the blue box, 1.5 million images are available. }
\label{fig:thailand-availability}
\end{figure*}

However, existing approaches remain either small in scale (i.e., rely on manual labeling) or are difficult to deploy in smallholder regions. 
Challenges in smallholder regions include: complex road networks in rural areas compared to the grid system present in the US Midwest \cite{yan2021exploring} and vegetation and man-made occlusions blocking the view between the road and fields (Figure \ref{fig:obstacles}).
Furthermore, as different regions grow different crops, current methods require hand-labelling a new crop type roadside dataset when encountering a new region.

In this paper, we propose a cost-effective automated deep learning pipeline to generate crop type references and remote sensing crop type maps 
with minimal manual labeling. We create a method to auto-generate field coordinates at scale, and then filter GSV images to create a large dataset of geo-tagged field images (Section 2.2). Next, a scraper automatically curates a training set of weakly-labeled images from either disparate out-of-domain sources or GPT-4V zero-shot classifications to train a CNN crop type discriminator on street-view images for the crop types of interest (Section 2.3). The crop type of field images is inferred by the CNN to generate ground referenced labels as a training set for a remote sensing crop type mapper (Sections 2.3-2.4). Once trained, the remote sensing model outputs crop type maps (Section 2.4). 

This work shows for the first time smallholder crop type mapping on a country scale using street-level imagery. We tested our approach in Thailand for the May 2022 to October 2022 wet growing season on the region's four major crops and created a ground truth set of 1400 GSV images labelled by a plant taxonomy expert. A total of 81,000 ground reference points were generated using our deep learning pipeline to train the remote sensing crop type model on the whole country. The crop type maps achieved an overall accuracy of 0.93 and an F1 score of 0.92.
The approach is orders of magnitude cheaper than traditional survey-based methods, scalable, high accuracy, and automated --- with expert hand-labeling only necessary to create a test set.

We publicly release the first-ever crop type map for all of Thailand 2022 with no gaps, 1400 geo-coordinates with crop types labeled by an agronomist, and 81,000 ground references. To our knowledge, this is the first time a 10m-resolution, multi-crop map (Figure \ref{fig:03-03}) has been created for any smallholder country. We open source all code, datasets, and crop type maps generated here: https://github.com/Earth-Intelligence-Lab/streetviewCropTypeMapping.

\section{Datasets \& Methods}
In brief, our method generates proposal locations along streets, curates a set of images containing crop fields (Section 2.2), predicts the crop type within the street-level image with minimal manual labeling (Section 2.3), and combines these labels with remote sensing time series to create a wall-to-wall crop type map (Section 2.4) (Figure \ref{fig:overview})(Figure \ref{fig:03-03}).

\subsection{Study Area} \label{msec:1}
We chose Thailand as the area of study, because it is simultaneously dominated by small-scale farms and has a high availability of Google Street View images. GSV images in Thailand grew 700\% from March 2022 to May 2023 to a total of 3.9M images, of which 2.8M are in cropland areas across the country (Figure \ref{fig:thailand-availability}). Two-thirds of cropland in the country is used to grow rice, while sugarcane, cassava, and maize are the next most abundant crops at just under 10\% of crop area each (Figure \ref{table:testSet}) \cite{faostat2021}. Rice is grown in two seasons, wet and dry; we limit this work to the wet season. 

\subsection{Finding Street-Level Images of Crop Fields} \label{msec:2}

\subsubsection{Extrapolate Equidistant Street Points} 

\begin{figure}[]
\centering
\includegraphics[width=0.48\textwidth]{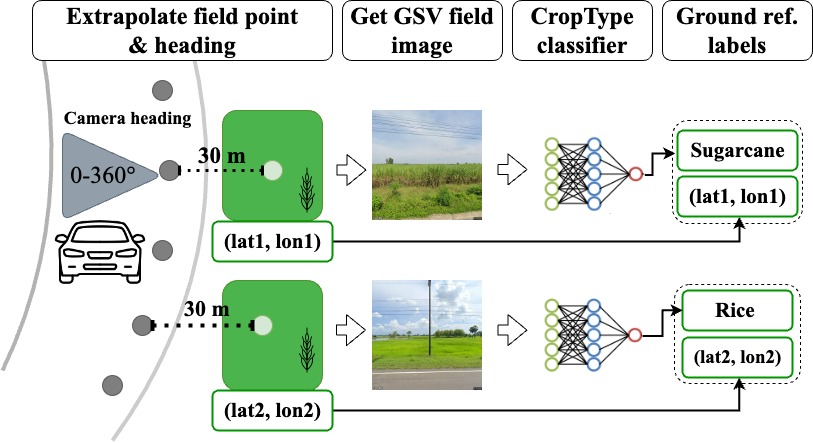}
\caption{Schematic of the process to generate ground reference labels. Each ground reference is composed of crop type and geocoordinates from street-view images.}
\label{fig:OSM-points}
\end{figure}

The first step in the pipeline is to gather the latitude and longitude coordinates of candidate fields and their corresponding roadside image along all roads in the country. We started with Open Street Map (OSM), a worldwide open geographic database with high coverage and completeness \cite{OpenStreetMap},
and used Overpass API to query all OSM ways within Thailand. In order to maximize recall of GSV field images, we generated equidistant points at 10m steps along OSM ways. Examples of points sampled from OSM street nodes in a chosen area are shown in Figure \ref{fig:OSM-points}.



\subsubsection{Filter Points Using Land Cover Map}

Next, we used existing land cover maps to filter for street points near farmland and remove field points that are not visible from the street due to obstructions (Figure \ref{fig:obstacles}). 
In particular, we used the European Space Agency's WorldCover 10m v100, which classifies global land cover at 10 meter resolution for the year 2020.
The map contains 11 land cover classes, including tree cover, shrubland, grassland, built-up, and cropland. We accessed WorldCover and filtered candidate points using functions available in Google Earth Engine.

Starting from the OSM-derived equidistant points, we removed points containing no cropland within a 10m radius. An inspection of 200 remaining points revealed that, in 30\% of GSV images, crop fields were still blocked by trees and other vegetation next to the road. We therefore also removed points containing any tree cover within the same 10m radius.
 

Finally, we are only interested in roadside images taken during the growing season, when crop types may be visible from the road. For this work, only GSV images captured during the wet growing season in Thailand (May 2022--October 2022) were considered. The above approach can be modified to any region and growing season worldwide by changing the date range and shapefile of the region of interest.

\subsubsection{Extrapolate Camera Heading and Field Point}

Due to the grid layout of streets in the US Midwest, prior work using GSV for crop type mapping obtained roadside images facing crop fields and points within crop fields by  extrapolating street points in north/south and east/west directions \cite{yan2021exploring}.
Since street layouts are more complex in smallholder regions, we first found the street bearing $\theta$ using the haversine formula (see Section A.2), and then computed the heading for the camera to face the two fields on either side of the street as $\theta \pm 90$ degrees (Figure 4).
We empirically determined that 30m was the best distance to extrapolate street points to crop field interior points (Table \ref{table:field_distance}).




\begin{table*}[t]
\small
\centering
\begin{tabular}{l r r r r r r r}
\toprule
& \multicolumn{4}{c}{\textbf{Percent of Dataset}} & \multicolumn{3}{c}{\textbf{Number of Samples}} \\
\cmidrule(lr){2-5}
\cmidrule(lr){6-8}
\multicolumn{1}{c}{\textbf{Crop Type}} & 
\multicolumn{1}{c}{\textbf{Planted Area}} &
\multicolumn{1}{c}{\textbf{Expert Street-view}} &
\multicolumn{1}{c}{\textbf{WebCC}} &
\multicolumn{1}{c}{\textbf{iNaturalist}} &
\multicolumn{1}{c}{\textbf{Expert Street-view}} &
\multicolumn{1}{c}{\textbf{WebCC}} &
\multicolumn{1}{c}{\textbf{iNaturalist}} \\
\midrule
Rice &  67\% & 66\% & 20\% & 14\% & 1261 & 659 & 1396 \\ 
Sugarcane  & 7\% & 7\% & 21\% & 20\% & 144 & 679 & 1882  \\ 
Cassava &  8\%  & 4\% & 18\% & 38\% & 81 & 584 & 3662\\ 
Maize &  8\%  & 6\% & 25\% & 18\% & 111 & 832 & 1698\\ 
Other & 10\% & 17\% & 16\%& 10\% & 325 & 512 &1008\\  
\midrule 
Total & 100\% & 100\% & 100\% & 100\% & 1922 & 3266 & 9646 \\
\bottomrule
\end{tabular}
\caption{Crop type distribution for the various datasets used to train and test a classifier on street-view images in Thailand. Planted area is obtained from annual national statistics released by the FAO. Expert street-view refers to GSV images randomly sampled from Thailand and manually labelled with crop type. WebCC are images scraped from the web. iNaturalist are images selected from an online biodiversity database.} 
\label{table:testSet}
\end{table*}

\subsubsection{Classify In-Field Images}

After filtering out points near trees and finding the appropriate camera heading, we found that 58\% of GSV images still had an obstructed view of a field  due to small bushes between the road and field not detected by satellite land cover maps (Table \ref{table:filteringGSV}).
We therefore labelled 2986 images into \{\textit{field}, \textit{not field}\} and trained a binary classifier using a ResNet-18 pre-trained on ImageNet. Hyperparameters used include an Adam optimizer and learning rate of $0.001$ for 15 epochs. The model classified field images from non-field images with 95\% recall and 98\% precision. 
Candidate street-view images were downloaded ($n = 224,000$) and classified by the \textit{field/not-field} CNN. Those labelled as \textit{field} ($n = 89,000$) were used as input in Section \ref{msec:3} to be classified into crop types. Although this work utilized 89,000 street-view images to generate ground references, there are approximately 800,000 GSV field images available in Thailand, offering substantial potential for the dataset to expand.


\subsection{Predicting Crop Type in Street-Level Images} \label{msec:3}

\paragraph{Compile Training Set from the Web} Data annotation to train a new classifier requires manual labeling, which is time consuming and, for crop type classification, requires domain expertise. Fortunately, the internet can serve as a source to rapidly obtain a training set of images that is large, low-cost, and diverse with real-world settings \cite{krause2016unreasonable}. 

We compiled training sets of crop type images from two sources: Google Images Creative Commons (hereafter ``WebCC'') and iNaturalist. For each crop type, images were queried in Google Images by searching for the crop name followed by ``field'' (e.g., ``rice field''). Returned images were labeled with the queried crop type. Meanwhile, iNaturalist is an online community of naturalists and citizen scientists who contribute photos of biodiversity across the globe. Its database contains over 161 million observations. While not targeted toward agriculture, iNaturalist contains images of crops, which can be downloaded via an API \cite{guillermo_huerta_ramos_2021_4733367}. In total, thousands of image-label pairs were collected from the two platforms (Table \ref{table:testSet}). 

A fifth ``Other" class was created to capture the remaining 10\% of crops planted in Thailand (by area). The class was made up of a random mix of other common crops grown in the country and downloaded in the same manner from the respective online source (e.g. ``barley field''). 

\begin{figure}[]
\centering
\includegraphics[width=0.45\textwidth]{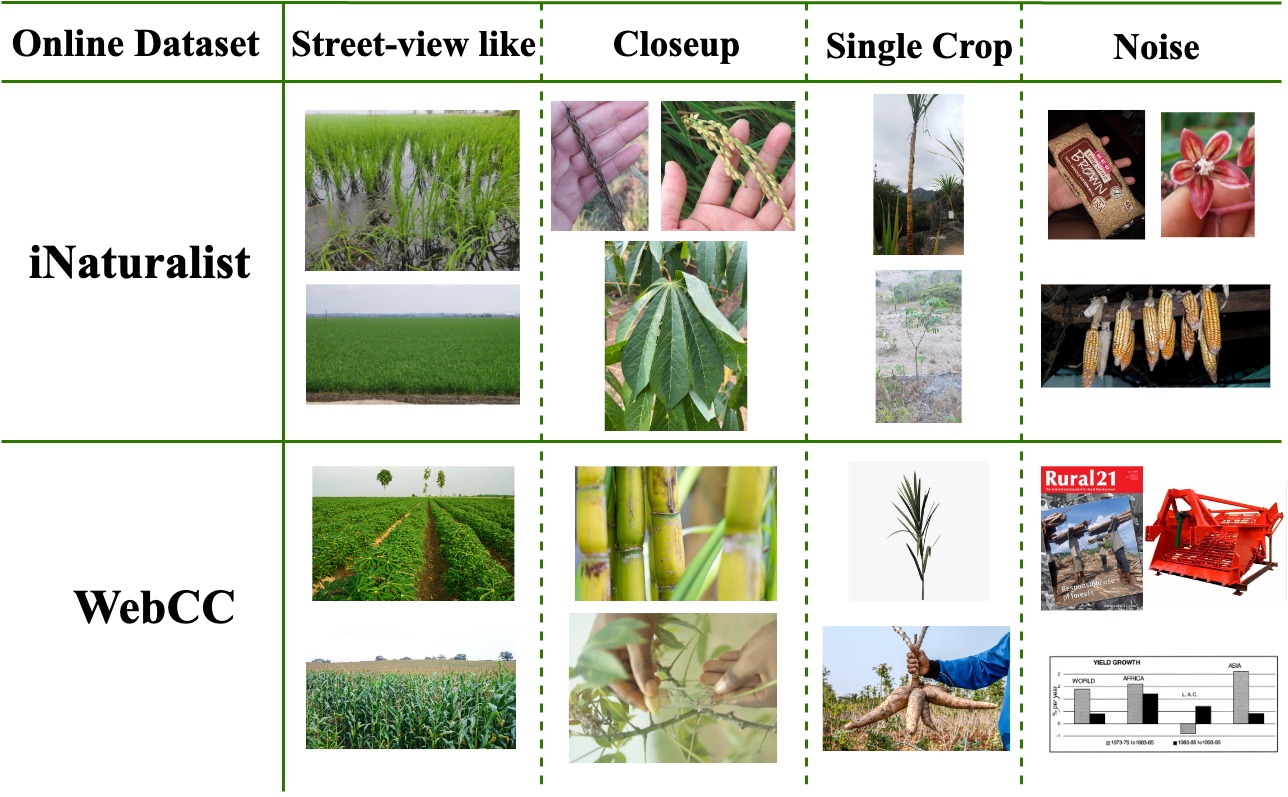}
\caption{Example images from the two online datasets. Although some images are of crop fields similar to the target street-view task, many images are either close-ups of a plant (especially in iNaturalist), a single plant instead of a field, or label noise (especially in WebCC).}
\label{fig:online-image-examples}
\end{figure}

\begin{figure}
\centering
\includegraphics[width=0.32\textwidth]{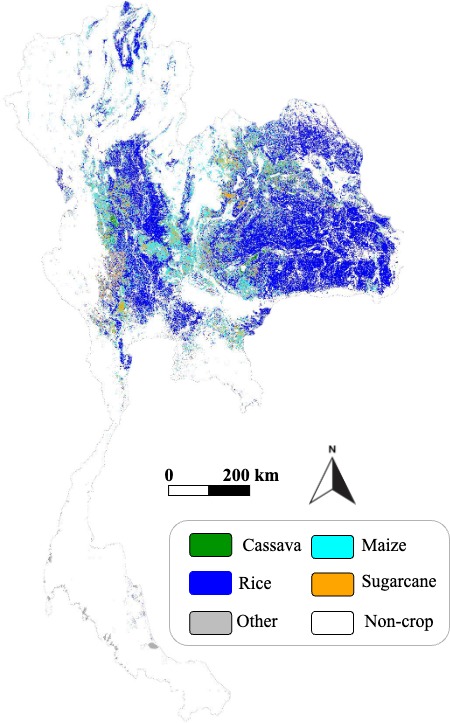}
 \caption{
      National crop type map of Thailand for 2022 at 10m resolution, created by combining satellite imagery and GSV crop type labels. The map has 93\% accuracy validated on 1,600 fields labeled by agronomists. We release this map publicly.
    } \label{fig:03-03}
\end{figure}

Although we tried to compile training images that look similar to GSV, online images varied in quality, size, and relevance. We observed many images of people with fields in the background, people holding crops, and label noise in WebCC (Figure \ref{fig:online-image-examples}). Fortunately, CNNs tend to be robust to noise and can perform well if enough signal from the desired task is present during training \cite{krause2016unreasonable, carlini2017towards, szegedy2013intriguing, NEURIPS2021_428fca9b}. Therefore, we did not manually clean the dataset.

We also explored the use of multimodal LLMs \cite{openai2023gpt4, gemini2023multimodal} to label a subset of GSV images and create a training set for a street-view classifier. Although these models have a small limit on queries per hour, their zero-shot performance could serve as an automated and cheaper method to create training set as compared to expert labeling. We used the prompt \textit{"What is the majority crop type in this image?"} and assigned the corresponding crop type label if it corresponded to one of the four main classes, otherwise we set it as ``Other". 

\paragraph{Manually Label Crop Type Ground References} We collected a test set of GSV images randomly sampled by geography and had them manually labeled by a plant taxonomy expert. To avoid data leakage in the remote sensing evaluation from two points belonging to the same field, we ensured all field points were more than 100m away from each other. We labelled 1922 images into the following classes: \textbf{cassava}, \textbf{maize}, \textbf{rice}, \textbf{sugarcane}, and \textbf{other}, the latter encompassing images tagged as unknown, non-crop, unsupported crop, and additional. The expert-labeled dataset served two purposes, (1) to compare performance of the CNN street-view classifier across different training sets and (2) to evaluate the remote sensing crop type classifier. 

\paragraph{Train a Street-Level Crop Type Classifier} We trained  models on 6 different datasets, 3 of which contained only online images automatically labeled by the corresponding search label. We use \textbf{WebCC} and \textbf{iNaturalist} to denote models trained on Google Images results and iNaturalist images, respectively. \textbf{iNaturalist + WebCC} was trained by combining the WebCC and iNaturalist datasets. \textbf{Expert labeled} was trained on 60\% of the expert-labeled GSV test set. \textbf{GPT-4V labeled} denotes a model trained using the same 1153 GSV images in the Expert labeled train set, but with labeling performed through GPT-4V's zero-shot classification. Finally, \textbf{Combined} merged the same Expert labeled train set of GSV with WebCC and iNaturalist. For all models, the remaining 40\% of the expert-labeled GSV dataset was split evenly into validation and test sets.

We used data augmentation techniques to accommodate images that varied in size across datasets and help the model focus on crop type features rather than a crop's position, zoom, or orientation in an image. During training, all images were expanded to $600 \times 600$ px and horizontally flipped at random, followed by a random crop of $300 \times 300$ px.

We selected a ResNet-50 architecture pre-trained on ImageNet for ease of reproducibility. All models were trained on 4-class classification with cross-entropy loss for 15 epochs, with a learning rate starting at $0.001$. A cosine annealing learning rate scheduler was used to dynamically adjust the learning rate and help improve model convergence. 

\subsubsection{Predict Crop Type in Street-View Images}

Trained models were used to classify street-view images ($n=89,000$) into crop types. We took sliding windows across each image, with a window of $300 \times 300$ pixels and a stride of $50$. 
We used the mode of high probabilities (MHP) to select the predicted class from all the sliding windows in an image. 
For each window, if $\textbf{p}$ is the vector of softmax probabilities and $c$ is the crop type with the highest probability, then the window casts one vote for class $c$ if $\textbf{p}_{c}$ exceeds some threshold $\tau$. If none of its probabilities exceed $\tau$, then the window casts no votes. The final prediction for the image is the crop type with the most votes across all windows.




Each clean street-view image was classified by the 5 models with sliding window and MHP and paired with its field coordinate to generate the ground reference training sets.

\begin{table}[t]
\centering
\small
\begin{tabular}{lc}
\toprule
\textbf{Pipeline Step (Section)} & \textbf{Dataset Size} \\
\midrule
Candidate field points (2.2) & 98 million \\ 
Filtered field points (2.2) & 3.9 million \\
GSV field images available (2.2) & 2.8 million\\
Street-view images downloaded (2.2) & 224,000 \\
Clean street-view images (2.2) & 89,000 \\
Labelled field points (2.3) & 89,000 \\
MHP thresholding (2.3) & 81,000 \\
\bottomrule
\end{tabular}
\caption{\textbf{Summary of dataset size through the point and image filtering steps in the pipeline}. Starting with a road network of Thailand and ending with 81,000 ground reference labels for the four major crop types. }
\label{tab:pipeline}
\end{table}
 
\subsection{Remote Sensing-Based Crop Type Mapping} \label{msec:4}

\subsubsection{Feature Extraction via Harmonic Regression}
Consistent with existing state-of-the-art methods \cite{inglada2015assessment, jin2019smallholder}, we used Sentinel-2 satellite time series for country-wide crop type classification. Sentinel-2 images 
capture 13 optical bands at 10-60m resolution on a 5-day cycle, and their time series contain information on crop phenology that allows different crop types to be distinguished.

\begin{table}[]
\small
\centering
\begin{tabular}{lcc}
\toprule
\textbf{Method} & \textbf{Precision} & \textbf{Recall}\\
\midrule
Baseline \cite{yan2021exploring} & 0.07 & 0.25 \\
Camera heading & 0.14 & 1.00 \\
Cropland filter & 0.31 & 1.00 \\
Cropland and tree cover filters & 0.42 & 0.99 \\
\midrule
All + field classifier & 0.98 & 0.95 \\
\bottomrule
\end{tabular}
\caption{Precision and recall for filtering out non-field GSV images. Baseline is our implementation of the GSV image collection approach in \citet{yan2021exploring}. Each \textbf{Method} includes the methods above. All methods previous to the field classifier are filters applied before downloading the GSV image.}
\label{table:filteringGSV}
\end{table}

\begin{table*}[h]
\small
\centering
\begin{tabular}{lrrrrrrr}
\toprule
& \multicolumn{6}{c}{\textbf{Street-view Test Set Metrics}}\\
\cmidrule(lr){2-8}
\textbf{Training Dataset} & \textbf{Overall Acc} & \textbf{Overall F1} & \textbf{Rice F1} & \textbf{Cassava F1} & \textbf{Maize F1} & \textbf{Sugarcane F1} & \textbf{Other F1} \\
\midrule
Baseline: Most common &  0.67 & 0.54 & 0.80 & 0.00 & 0.00 & 0.00 & 0.00 \\ 
\midrule
WebCC & 0.82 $\pm$ 0.04& 0.82 $\pm$ 0.03 & 0.90 $\pm$ 0.03 & 0.73 $\pm$ 0.07 & 0.62 $\pm$ 0.09 & 0.62 $\pm$ 0.04 & 0.49 $\pm$ 0.07 \\
iNaturalist & 0.81 $\pm$ 0.04 &0.82 $\pm$ 0.03 & 0.89 $\pm$ 0.02 &  0.63 $\pm$ 0.06 & 0.68 $\pm$ 0.06  & 0.76 $\pm$ 0.04 & 0.51 $\pm$ 0.08 \\
iNaturalist + WebCC & 0.85 $\pm$ 0.02 & 0.85 $\pm$ 0.02 & 0.91 $\pm$ 0.02& 0.75 $\pm$ 0.04& 0.71 $\pm$ 0.06 & 0.78 $\pm$ 0.03 & 0.62 $\pm$ 0.06\\
\midrule
\textbf{GPT-4V (zero-shot)\textsuperscript{\textdagger}} & \textbf{0.95 $\pm$ 0.00} & \textbf{0.95 $\pm$ 0.00} & \textbf{0.97 $\pm$ 0.00} & \textbf{0.97 $\pm$ 0.00} & \textbf{0.89 $\pm$ 0.00} & \textbf{0.93 $\pm$ 0.00} &  \textbf{0.87 $\pm$ 0.00}\\
GPT-4V labeled & 0.92 $\pm$ 0.01 & 0.92 $\pm$ 0.01 & 0.94 $\pm$ 0.01& 0.86 $\pm$ 0.03 & 0.81 $\pm$ 0.02 & 0.90 $\pm$ 0.02 & 0.77 $\pm$ 0.06\\
\midrule
\textbf{Expert labeled} & \textbf{0.93 $\pm$ 0.01} & \textbf{0.93 $\pm$ 0.01} & 0.94 $\pm$ 0.01 & \textbf{0.87 $\pm$ 0.03} & \textbf{0.82 $\pm$ 0.03} &   \textbf{0.92 $\pm$ 0.02} & \textbf{0.79 $\pm$ 0.05}\\
Combined & 0.93 $\pm$ 0.01 & 0.93 $\pm$ 0.01 & \textbf{0.95 $\pm$ 0.01} & 0.86 $\pm$ 0.03 & 0.81 $\pm$ 0.03 & 0.92 $\pm$ 0.02 & 0.77 $\pm$ 0.05\\
\bottomrule
\end{tabular}

\caption{Performance on crop type classification from street-view images for the four major crops in Thailand. Models were trained on combinations of three different training datasets: WebCC, iNaturalist, GPT-4V labeled, and Expert labeled GSV images. The MHP threshold was set to 0.9. \textsuperscript{\textdagger}GPT-4V refers to GPT-4V zero-shot performance on the test set. Each model is trained 5 times with different seeds and bootstrapped training sets. Datasets are ordered by increasing labeling cost.}
\label{table:gsv-results}
\end{table*}

We used Google Earth Engine to export Sentinel-2 L2A time series in Thailand from May 1 -- October 31, 2022. Four spectral bands were used: Red Edge 4, SWIR 1, SWIR 2, and NIR. We added the green chlorophyll vegetation index ($\text{GCVI} = \text{NIR}/\text{GREEN} - 1$), as prior work showed it to be a valuable feature for crop type classiﬁcation. We used the Cloud Probability band to remove cloudy days before extracting time-series for each ground reference.

We used the harmonic regression to extract frequency-domain features from time series of varying lengths for input to machine learning (Figure \ref{fig:time-series-extract}). Equivalent to the discrete Fourier Transform, harmonic regression generates robust features for crop type classification \cite{shumway2000time,jin2019smallholder, ghazaryan2018rule, jakubauskas2001harmonic, wang2019crop, wang2020mapping}. We applied a 3\textsuperscript{rd} order harmonic regression to each band independently to arrive at a total of 35 features. 




\subsubsection{Country-Wide Crop Type Classification}
We trained random forests (with $500$ trees) on crop type classification, with the harmonic coefficients as input and the crop type labels generated by the street-view CNN as output. Random forests \cite{breiman2001random} are frequently used in remote sensing applications for their high accuracy and computational efficiency \cite{gislason2006random, azzari2017landsat, ok2012evaluation}.

The crop type ground references ($n = 81,000$, Section 2.3) were used to train the remote sensing crop type classifier. We trained 5 random forests, one for each ground reference set (Table \ref{table:gsv-results}). A sixth model was trained only on the expert-labeled dataset ($n=1,153$). Compared to automatically-labeled points, expert-labeled points are more accurate but almost 100 times smaller in quantity.

\section{Results}
\subsection{Automatically Generated Field Points and Street-View Images} \label{rsec:2}

We obtained 98 million points across Thailand by computing 10m equidistant points from OSM ways.  Of these points, 3.9 million occurred near cropland without tree cover, and 2.8 million had GSV images available (Table \ref{tab:pipeline}). Of these street-view images, we downloaded 224,000 and classified 89,000 into images of fields and 135,000 into not fields. The 89,000 field-view images were classified by crop type (rice, maize, cassava, sugarcane) and passed though the MHP threshold to create 81,000 point ground reference labels. The remaining 8,000 images were removed at MHP thresholding because none of their sliding window predictions had a softmax probability greater than 0.9 indicating label uncertainty.  

To understand how important land cover filtering, camera heading, and \textit{field/not field} classification were for finding GSV images of crops, we calculated the precision and recall at various steps along the pipeline (Table \ref{table:filteringGSV}). We also implemented the method developed by \citet{yan2021exploring} in the US Midwest, which did not include these filtering steps, as a baseline. In a sample of random GSV images in Thailand, we found that the baseline method achieves a precision of 0.07 and recall of 0.25; in other words, 93\% of downloaded GSV images did not show a crop field, and 75\% of images of crop fields were missed. This illustrates the complexity of smallholder landscapes compared to industrial agriculture. By contrast, using the correct camera heading on points generated 10m apart from OSM ways yielded a precision of 0.14 and recall of 1.00. Filtering by cropland alone improved precision to 0.31, and removing points near trees further improved precision to 0.42. Finally, training a CNN to classify \textit{field/not field} improved precision to 0.98 and lowered recall slightly to 0.95 --- a worthwhile trade-off.



\subsection{Weakly Supervised and AI-Assisted Creation of Crop Type Ground References} \label{rsec:2}

\begin{table*}[h]
\small
\centering
\begin{tabular}{cccccccc}
\toprule
& \multicolumn{7}{c}{\textbf{Remote Sensing Test Set Metrics}}\\
\cmidrule(lr){2-8}
\textbf{Ground Ref. Train Data} & \textbf{Overall Acc} & \textbf{Overall F1} & \textbf{Rice F1} & \textbf{Cassava F1} & \textbf{Maize F1} & \textbf{Sugarcane F1} & \textbf{Other F1}\\
\midrule
Baseline\textsuperscript{*} &  0.65 & 0.51 & 0.79 & 0.00 & 0.00& 0.00 & 0.00 \\ 
Auto w/ WebCC\textsuperscript{\textdagger} &0.80 $\pm$ 0.01 & 0.80 $\pm$ 0.01 & 0.89 $\pm$ 0.02 & 0.62 $\pm$ 0.07 & 0.60  $\pm$ 0.08  & 0.81 $\pm$ 0.01 & 0.51 $\pm$ 0.05 \\
Auto w/ iNaturalist\textsuperscript{\textdagger}&  0.78 $\pm$ 0.02& 0.77 $\pm$ 0.02 & 0.88 $\pm$ 0.03 & 0.61 $\pm$ 0.05 & 0.67 $\pm$ 0.04 & 0.70 $\pm$ 0.03 & 0.53 $\pm$ 0.05  \\
Auto w/ iNat+WebCC\textsuperscript{\textdagger} &  0.82 $\pm$ 0.01  & 0.82 $\pm$ 0.02  & 0.90 $\pm$ 0.02 & 0.68 $\pm$ 0.05 & 0.69  $\pm$ 0.03 & 0.81 $\pm$ 0.02 & 0.55 $\pm$ 0.05 \\
\midrule
Auto w/ GPT-4V GSV\textsuperscript{\textdagger} &  0.92 $\pm$ 0.02 & 0.92$\pm$ 0.02 & 0.95 $\pm$ 0.02 & 0.87 $\pm$ 0.04 & 0.74 $\pm$ 0.02& 0.90 $\pm$ 0.02 & 0.87 $\pm$ 0.02\\
\midrule
Auto w/ Expert GSV\textsuperscript{\textdagger} & \textbf{0.93 $\pm$ 0.01} & \textbf{0.93 $\pm$ 0.01} & \textbf{0.95 $\pm$ 0.01} & \textbf{0.89 $\pm$ 0.04} & \textbf{0.75 $\pm$ 0.03}  & \textbf{0.91 $\pm$ 0.01} & \textbf{0.91 $\pm$ 0.02}\\
Auto w/ Combined\textsuperscript{\textdagger} &  0.92 $\pm$ 0.02 & 0.92 $\pm$ 0.02 & 0.95 $\pm$ 0.02& 0.87 $\pm$ 0.04 & 0.74 $\pm$ 0.03& 0.91 $\pm$ 0.02& 0.88 $\pm$ 0.03\\
\midrule
Expert GSV\textsuperscript{\textdaggerdbl}& 0.69 $\pm$ 0.04 & 0.69 $\pm$ 0.04  & 0.82 $\pm$ 0.05& 0.27 $\pm$ 0.08& 0.23 $\pm$ 0.08& 0.54 $\pm$ 0.05& 0.61 $\pm$ 0.09\\
\bottomrule
\end{tabular}
\caption{Performance on remote sensing-based crop type classification for the four major crop types in Thailand.\textsuperscript{*}Baseline refers to classifying all samples as the most common class, rice.
\textsuperscript{\textdagger}Ground references are automatically labeled using CNNs trained on the specified datasets ($n = 81,000$). \textsuperscript{\textdaggerdbl}Ground references contain only expert-labeled GSV images ($n = 984$). Each model is trained 5 times with different seeds and bootstrapped training sets and then ran through the rest of the pipeline to obtain final error bars. Datasets are ordered by increasing labeling cost.}
\label{table:remote-results}
\end{table*}

\subsubsection{Street-View Crop Type Classification}
Our experiments on the 6 datasets showed that weak supervision with online images can successfully classify crop type in street-view images (Table \ref{table:gsv-results}). WebCC ($n = 3,266$) on its own achieved 82\% overall accuracy, similar to iNaturalist ($n = 9,646$) at 81\%. For comparison, a baseline that classifies all samples as the most common class (rice) would achieve 67\% accuracy. Maize had the lowest F1 score under both WebCC and iNaturalist (62\% and 68\%, respectively), followed by sugarcane (62\% and 76\%). The low performance is likely due to the more visual similarities between the two crops at the early-growth stage as well as their similar height. Meanwhile, rice has short and bright green leaves and cassava has a  palmate leaf structure and is grown more separately, both more visually distinctive from the street. 

When merged together, iNaturalist + WebCC ($n = 12,912$) overall accuracy improved to 85\%. F1 scores for cassava, maize, sugarcane and other also substantially improved. This complementarity could be due to iNaturalist having higher label accuracy but more out-of-domain images (e.g. closeups), while WebCC has lower label accuracy but considerably more street-view-like images. 

GPT-4V demonstrated consistently high zero-shot classification performance in all categories, with the highest accuracy (95\%). Upon inspecting its responses, it showed a clear understanding of the distinctive phenotypes of the plants to support its decision. 

The CNN trained on the expert-labeled dataset ($n = 1,153$) achieved 93\% accuracy, nearly identical to the Combined model ($n = 12,339, 93\%$ accuracy).  The comparable performance of the CNN trained on GPT-4V labels ($n = 1,153)$ shows that today's multimodal LLMs can already help minimize the need for expert labeling.

\subsubsection{Role of Sliding Window Classification and Probability Threshold}

The MHP approach to remove low-confidence classifications (Section \ref{msec:3}) proved to increase the accuracy and F1 score across all models for thresholds set to 0.7 or above. The top performing model trained on the Expert labeled dataset achieved 82\% accuracy when directly classifying the whole image, 90\% with sliding windows but no MHP threshold, and 93\% with a 0.90 MHP threshold. 
Improvements were similar for the other training sets. 

We note that a higher MHP threshold $\tau$ leads to some images being dropped from the inference set, because sometimes no sliding windows have softmax probabilities exceeding $\tau$. Despite the slight loss of data, we observed that F1 score increased monotonically with $\tau$ all the way up to $\tau=0.95$.



\subsection{Country-Wide Crop Type Map}


\subsubsection{Automated vs. Expert-Labeled GSV Ground References}
We found that training a random forest on large CNN-generated GSV ground references ($n = 81,000$) resulted in crop type maps that were significantly more accurate than training on small expert-labeled ground references ($n = 984$), despite the CNN-based labels being imperfect (Table \ref{table:remote-results}).
Training on expert-labeled ground references achieved 69\% overall accuracy, which is modest given the 65\% accuracy of a baseline model that classifies everything as the most common class (rice). In comparison, the random forest trained on 81,000 GSV samples whose crop types were predicted by the street-view CNN trained on expert labels achieved an accuracy of 93\%. This suggests that the sample volume generated by leveraging GSV outweighs the noise that predicted crop type labels can carry from misclassification.

\subsubsection{Weakly Supervised and AI-Assisted vs. Expert-Labeled Automated GSV Ground References}
Among the 6 automatically-generated GSV ground reference datasets, the datasets generated by CNNs trained on expert labels or GPT-4V labels proved more accurate than those generated by CNNs trained on weak labels from the web.
Consistent with the street-view classification results in Table \ref{table:gsv-results}, the lowest-performing datasets were those created by the WebCC CNN and the iNaturalist CNN; the remote sensing-based crop type classifier trained using their predictions as labels achieved accuracies of 80\% and 78\%, respectively.
The crop type classifer trained on iNaturalist + WebCC CNN predictions performed better at 82\% accuracy, but suffered from low cassava and maize F1 scores (68\% and 69\%). The three highest-performing crop type classifiers were trained on outputs of the GPT-4V, expert-labeled, and Combined CNNs, achieving accuracies of 92\%, 93\%, and 92\% and higher F1 scores across all five crop classes. 


For the individual crop types, rice---the most abundant crop---was classified most accurately, with F1 scores of 95\% for the top 3 models. Sugarcane was also classified accurately, with F1 scores ranging from 90\% to 91\% for the top 3 models. Maize and cassava were consistently the most difficult crop types to classify across all models.

The weakly-supervised CNNs struggled to classify the ``Other" class, but the GPT-4V and expert-labeled models performed better. We observed in experiments that including the ``Other" class  enhanced the performance of the four main crop type classes. 

\subsubsection{Sensitivity to Training Set Size}
\begin{figure}[t]
\centering
\includegraphics[width=0.47\textwidth]{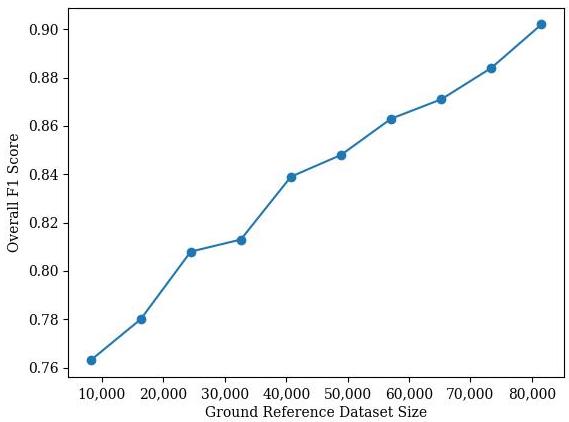}
\caption{Crop type map F1 score vs. ground reference dataset size. Computed for the Auto w/ iNat + WebCC ground reference set from Table \ref{table:remote-results}.}
\label{fig:f1-vs-trainsize}
\end{figure}

Lastly, we investigated the relationship between the number of ground references and the remote sensing-based crop type classification performance. We found that, even at 81,000 ground references, performance had not yet saturated as a function of training set size; the overall F1 score continued to increase linearly as more street-view points were added (Figure \ref{fig:f1-vs-trainsize}).


\section{Discussion}
We show that deep learning and street-view images can be combined to generate thousands of geolocated crop type ground references at scale in smallholder regions. These ground references, despite containing some label noise, can then be used to create high-accuracy crop type maps in countries where no such maps currently exist.
In Thailand, 81,000 automated ground references led to a more accurate crop type map than 1000 expert-labeled ground references, and even at 81,000 performance had not yet saturated as a function of sample size.

To minimize the need for experts to manually label crop types in street-view images, we explored using images from the web to weakly supervise a CNN to classify crop type. We found that images from Google Images and iNaturalist, although high in noise and often off-domain, can successfully supervise the classification of street-view images.
Furthermore, combining images from different online sources improved performance. We also found that a CNN trained on labels generated by GPT-4V zero-shot classification showed comparable results to the model trained on expert labels.
When creating the country-wide crop type map in Thailand, we did observe that weakly supervised CNNs led to lower accuracy crop type maps than CNNs trained on Expert or GPT-4V labels, although that difference disappears when weak labels are combined with expert labels. At the time of writing, the best trade-off between cost of labeling and accuracy of the final crop type map appears to be to use GPT-4V to label roadside imagery. However, we note that we only validated GPT-4V's ability for the four main crops in this paper. We also highlight that current GPT-4V query limits make it difficult for most users to label more than a few thousand images per month. 



Limitations of our work include that, to train a classifier to remove street-view images containing small bushes, we manually labeled a set of \textit{field/not field} images. However, we point out that, unlike crop type labeling, \textit{field/not field} labeling does not require domain expertise;  furthermore, this step could also potentially be classified by GPT-4V.
Another limitation
is the uncertain update frequency of street-view services and their continued uneven distribution around the globe. 
For these reasons, we do not see GSV as the one-fix-all solution for global crop type mapping. However, GSV does cover 98 countries, of which only 23 have crop type maps, and it updates every $\sim$3 years. GSV is also still expanding; in 2022 it launched new lightweight cameras mapping India and parts of Latin America. We also note that our approach extends to non-GSV street-level imagery as well. Lastly, remote sensing models may transfer geographically once trained on sufficient data \cite{beery2022auto}, so models trained in GSV regions could be used in non-GSV regions as long as test data exists locally to evaluate performance. Models could also transfer over time; we leave such explorations to future work. 


We release all datasets used for training and testing each model, including over 81,000 crop type ground reference points, and the code to run this pipeline in other regions. We also release the crop type map created for Thailand's 2022 wet growing season, which currently does not exist. 

\appendix

\section{Appendices}

\subsection{Generating equidistant points from OSM}

Line segments are created for each OSM way, by creating an edge between OSM node pairs using linear interpolation. Then starting with the point at the base of the line, latitude and longitude points are generated at 10 meter steps recursively over the $\text{distance}$ between the node pairs as shown below:

\begin{equation}
\text{lat}_{\text{new}} = \text{lat}_{\text{prev}} + \frac{\text{distance}}{10 \cdot (\text{lat}_\text{new} - \text{lat}_\text{prev})}
\end{equation}

\begin{equation}
\text{lon}_{\text{new}} = \text{lon}_{\text{prev}} + \frac{\text{distance}}{10 \cdot (\text{lon}_\text{new} - \text{lon}_\text{prev})}
\end{equation}

\begin{figure}[b]
\centering
\includegraphics[width=0.35\textwidth]{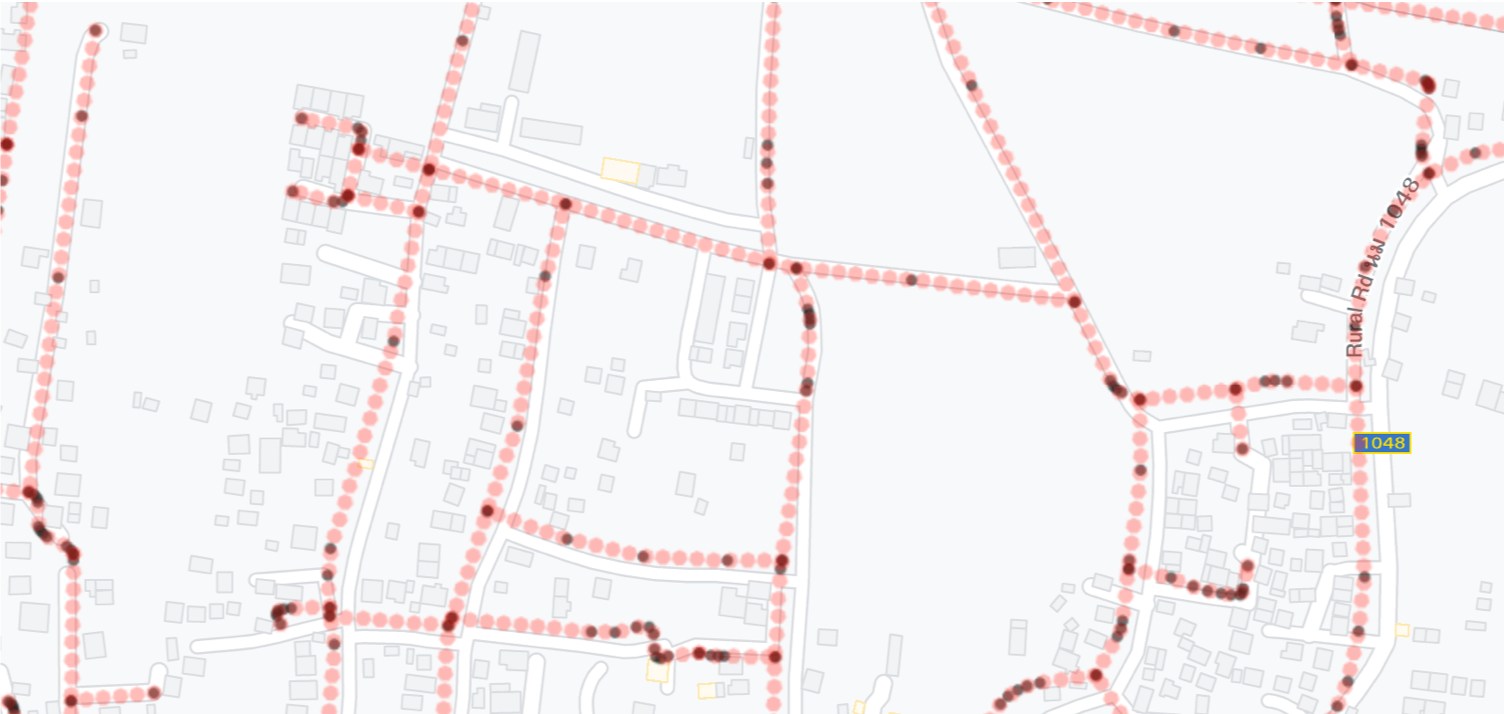}
\caption{OSM Points in black and equidistant generated points in red. OSM points gathered from OSM ways are random and sparse. The red points are generated at 10m steps in order to query GSV and ensure higher recall of field points and street-view images.    }
\label{fig:osm-points}
\end{figure}

\subsection{Pulling GSV street-level imagery}
Google Street View (GSV) is used as the dataset to gather images at the desired geocoordinates in fields. To query the GSV API, the arguments (lat, lon) and camera heading ($\theta$) are used to download images at the desired location with the camera facing towards the field. GSV charges \$7 for 1000 images, which is why a significant filtering on the set of points is performed beforehand using land cover maps.
The street bearing is necessary for two purposes (1) to pull GSV images with the camera facing the direction of the field, that is, +90 degrees and -90 degrees from the street bearing and (2) to calculate the point coordinate of labelled fields that will be fed to the remote sensing model.

To achieve this the road bearing for each filtered street point is interpolated using Harversine formulas - which account for earth's ellipsoidal shape. The formulas calculate the direction $\theta$ from a start point $(\text{lat}_1, \text{lon}_1)$ to its adjacent point $(\text{lat}_2, \text{lon}_2)$ 10m away. Latitudes and longitudes are in radians.

\begin{equation}
\Delta lat = lat_2 - lat_1
\end{equation}
\begin{equation}
y = \operatorname{sin}(lat_2 - lat_1) \cdot \operatorname{cos}(lon_2) 
\end{equation}
\begin{equation}
x = \operatorname{cos}(lon_1) \cdot \operatorname{sin}(lon_2) - \operatorname{sin}(lon_1) \cdot \operatorname{cos}(lon_2) \cdot \operatorname{cos}(\Delta lat) 
\end{equation}
\begin{equation}
\theta = \operatorname{atan}\tfrac{y}{x}
\end{equation}

Once the street bearing is calculated and converted back to degrees, the bearing for the camera to face the two fields on either side of the street are $\theta$ + 90 and $\theta$ - 90.

Next we calculate the field point D meters away, perpendicular to the street bearing, where a GSV image was taken. This field point will be labelled in section 4.2 and used as ground reference in training data for remote sensing in section 4.3. To achieve this the direct haversine formula is used.
$\textit{lat}_s$ and $\textit{lon}_s$ are street point coordinates to calculate, $\textit{lat}_f$ and $\textit{lon}_f$ are field point coordinates, \textit{D} is the distance away from the street, $\theta$ is the direction, \textit{R} is the mean radius of the earth, and $\textit{A}_D$ = $\frac{D}{R}$. 

\begin{equation}
\begin{split}
lon_f &= \operatorname{asin}(\sin(lon_s) \cdot \cos(A_D)  \\
&\quad+ \cos(lon_s) \cdot \sin(A_d) \cdot \cos(\theta)) 
\end{split}
\end{equation}

\begin{equation}
lat_f = lat_s + \operatorname{atan}(\tfrac{\sin(\theta) \cdot \operatorname{sin}(A_d) \cdot \operatorname{cos}(lon_s)} {\operatorname{cos}(A_D) - \operatorname{sin(lon_s)} \cdot \operatorname{sin}(lon_f)})
\end{equation}



The $(lon_{f}, lat_{f})$ field geocoordinates are assigned to their respective image to be labelled, to serve together as a ground reference point to train the remote sensing models.

\begin{table}[h]
\small
\centering
\begin{tabular}{ccc}
\toprule
\textbf{Distance (m)} & \textbf{Furthest FB\%} & \textbf{Closest FB\%} \\
\midrule
\textless 10 & 0 & 8 \\
10 & 0 & 83 \\
20 & 0 & 9 \\
30 & 2 & 0 \\
40 & 21 & 0 \\
50 & 41 & 0 \\
\textgreater 50 & 36 & 0 \\
\bottomrule
\end{tabular}

\caption{Showing the distance of furthest field boundary from to the street, and the distance of the closest boundary from the street. The data was collected by inspecting random samples across Thailand for 200 fields. FB stands for field boundary. The experiment showed placing the ground reference 30 m perpendicular from where the GSV image was taken is reasonable with only 2\% of outliers. }
\label{table:field_distance}
\end{table}

\begin{figure}[]
\centering
\includegraphics[width=0.46\textwidth]{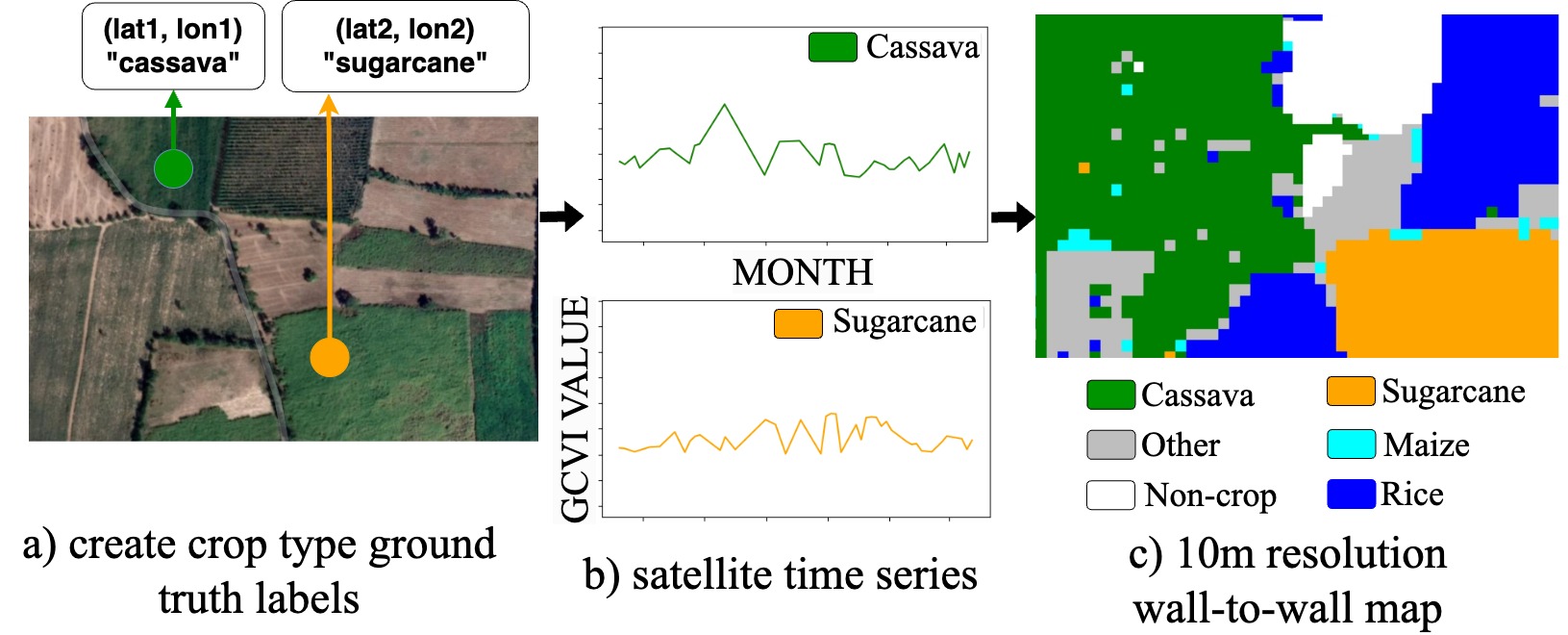}
\caption{Extracting satellite time series from
crop type ground labels to train the national crop type map.}
\label{fig:time-series-extract}
\end{figure}

\bibliography{aaai24.bib}

\end{document}